\documentclass[letterpaper,journal]{IEEEtran}

\usepackage[T1]{fontenc}
\usepackage{amsmath,amsfonts,amssymb}
\usepackage{array}
\usepackage{multirow}
\usepackage{booktabs}
\usepackage[caption=false,font=footnotesize]{subfig}
\usepackage{textcomp}
\usepackage{stfloats}
\usepackage{placeins}
\usepackage{url}
\usepackage{graphicx}
\usepackage{xcolor}
\usepackage{cite}
\usepackage[hidelinks]{hyperref}
\usepackage[capitalise]{cleveref}

\graphicspath{{figs/}{figures/}{pictures/}{images/}{./}}
\begin{document}

\title{Instance Segmentation and Fine-grained Classification for Urban Buildings with Adaptive Region Dividing and Spatially-Supervised Contrastive Learning}

\author{Weiyuan Zhang, 
Qi Zhang*\thanks{* Corresponding author: Qi Zhang.}, 
and Hui Huang%
\thanks{Weiyuan Zhang, Qi Zhang, and Hui Huang are with Guangdong Provincial Key Laboratory of Visual Media and Multidimensional Intelligence, College of Computer Science and Software Engineering, Shenzhen University. E-mail: 2400101002@mails.szu.edu.cn, qi.zhang.opt@gmail.com, and hhzhiyan@gmail.com.}%
}

\maketitle

\begin{abstract}
Accurate instance-level and functional understanding of urban buildings in large-scale point clouds is essential for digital city modeling and urban analysis. However, the extensive spatial coverage of urban scenes leads most existing methods to rely on predefined blocks for training and evaluation, although such partitions are rarely available in real-world applications and introduce additional preprocessing while fragmenting complete building structures. To address this issue, we propose an adaptive region-dividing strategy with unified scene-level evaluation. Specifically, the 3D point cloud is projected onto a bird's-eye-view (BEV) plane, where a pretrained segmentation model is used to detect building regions. The detected bounding boxes are then back-projected to the original point cloud to construct structure-aligned adaptive training blocks, enabling semantically guided dynamic partitioning without manual design. Furthermore, beyond instance-level understanding, few methods have explored fine-grained classification for urban buildings, and thus we also put forward a fine-grained classification model for urban buildings with a spatially-supervised contrastive loss. First, for each segmented building, a point transformer classifier jointly encodes its body and local context using geometric, color, and core-context information. Then, the class-balanced weighted cross-entropy is used to alleviate severe class imbalance. The proposed spatially-supervised contrastive loss further enhances inter-class discriminability by assigning greater weight to spatially proximate, same-category buildings, encouraging compact functional representations while separating easily confused categories. Extensive experiments on UrbanBIS and STPLS3D demonstrate the advantages of the proposed method in building instance segmentation and fine-grained classification compared to existing SOTA methods.
\end{abstract}


\section{Introduction}
\IEEEPARstart{L}{arge-scale} aerial photogrammetric urban point clouds \cite{Yang2023UrbanBIS,Chen2022STPLS3D} capture building geometry, appearance, and spatial organization, supporting urban mapping, planning, and infrastructure management. Identifying individual buildings and their functions extends object separation to semantic interpretation. However, extensive spatial coverage, high building density, and heterogeneous structures make complete building-level understanding challenging.

From a multimedia understanding perspective, urban point clouds are 3D visual content requiring both coherent object structure and discriminative semantic representations. Complete objects connect geometry and appearance with functional meaning, while contrastive learning organizes feature relationships beyond individual class predictions \cite{Wu2024IntraCrossModal}. Related studies explore supervised contrastive learning for indoor oversegmentation \cite{Sun2025Oversegmentation} and adaptive-margin contrastive learning for ambiguity-aware segmentation \cite{Chen2026Ambiguity}. These directions motivate using image-space semantics to organize 3D observations and spatial context to improve building representations.

\begin{figure}[!t]
    \centering
    \includegraphics[width=\columnwidth]{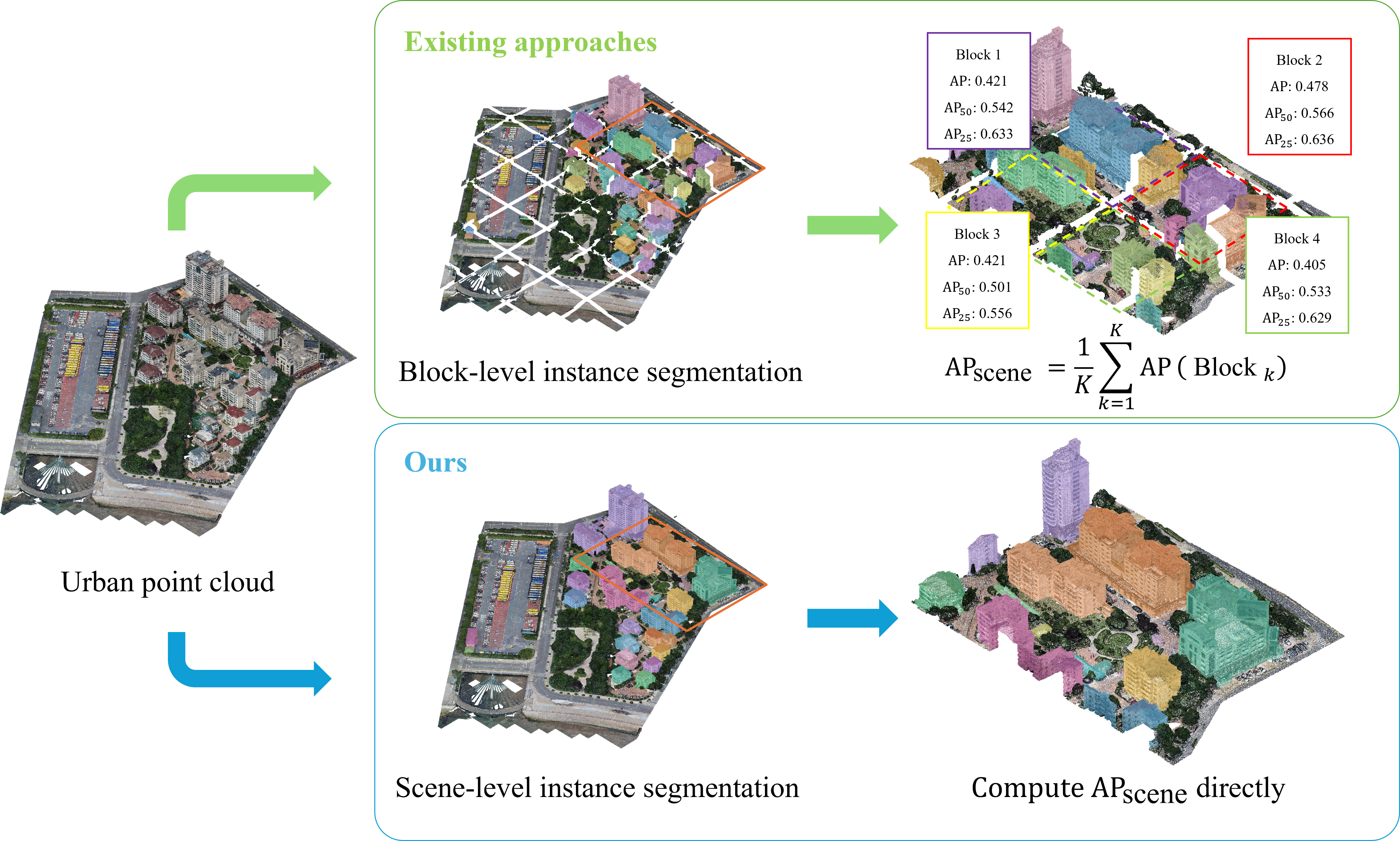}
    \vspace{-0.5cm}
    \caption{Comparison between conventional block-level and our scene-level evaluation for urban building instance segmentation. For the same test scene, existing approaches divide the point cloud into local blocks and predict and evaluate instances within each block, which may fragment buildings crossing block boundaries. In contrast, our method directly processes the unpartitioned test scene and evaluates instances in the original scene space, avoiding partition-induced fragmentation and producing a unified result.}
    \label{fig:teaser}
    \vspace{-0.5cm}
\end{figure}

\begin{figure*}[!t]
    \centering
    \includegraphics[width=\textwidth]{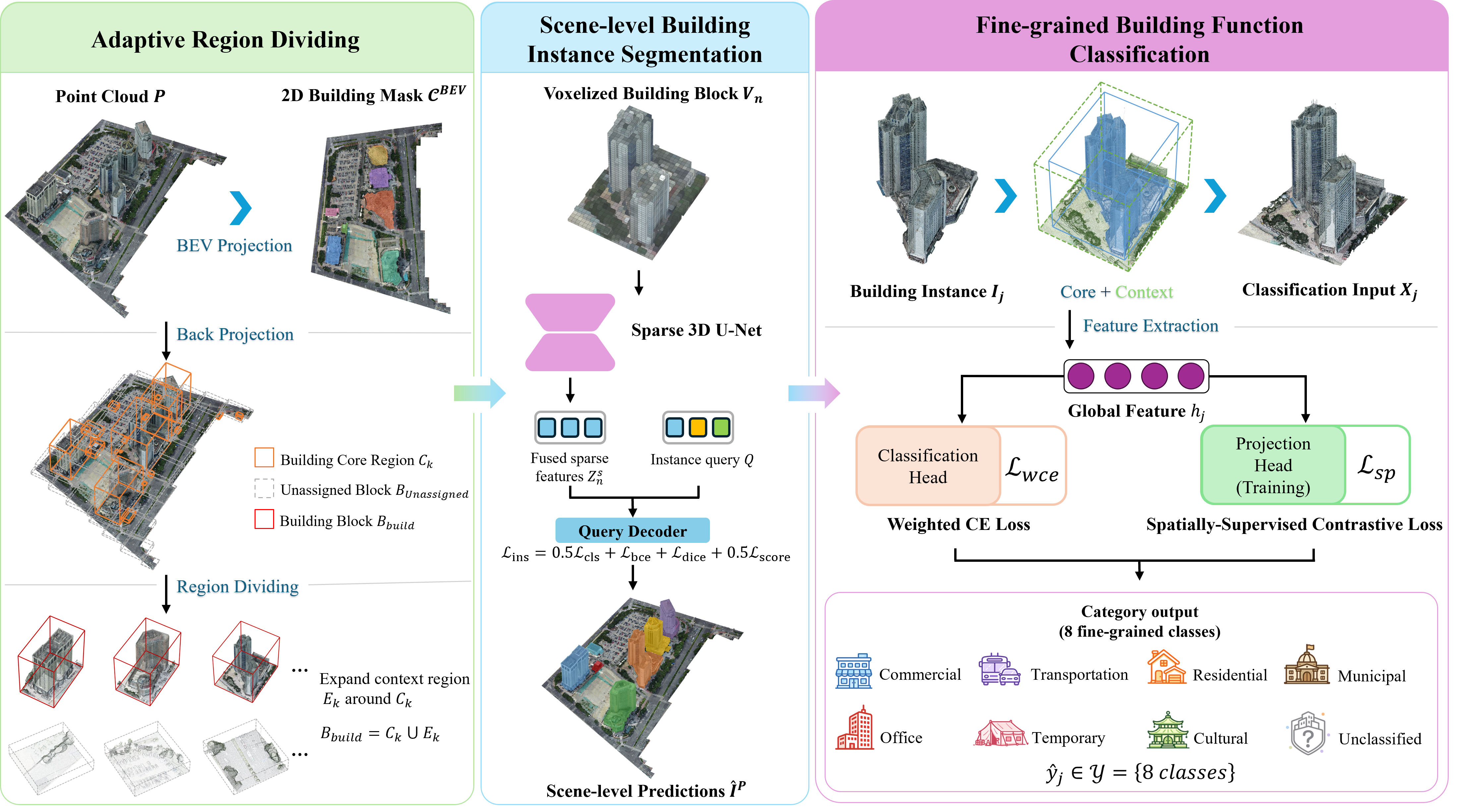}
    \vspace{-0.4cm}
    \caption{Overview of the proposed framework. Adaptive region dividing projects the point cloud into BEV space, constructs building-aware blocks from core regions and expanded context, and partitions unassigned points into regular blocks. These adaptive blocks train the instance segmentation network, whereas the unpartitioned scene is directly processed during scene-level evaluation. A sparse 3D U-Net extracts multi-scale voxel features, and a query decoder predicts building instance masks that are mapped to the original points. For fine-grained classification, each building is combined with optional local context and encoded by PTv3. The classification head predicts one of eight building-function categories, whereas the projection head is used only during training for spatially-supervised contrastive learning.}
    \label{fig:pipeline}
    \vspace{-0.2cm}
\end{figure*}

Existing instance segmentation networks commonly use room-scale or bounded inputs \cite{Jiang2020PointGroup,Chen2021HAIS,Vu2022SoftGroup,Schult2023Mask3D,Kolodiazhnyi2024OneFormer3D}. Urban pipelines therefore divide scenes into predefined blocks to control memory and computation \cite{Yang2023UrbanBIS}. These blocks can fragment buildings and discard context, making supervision dependent on partition placement. Large or irregular buildings are particularly affected because fixed boundaries can separate their structural components. Recovering scene predictions requires duplicate suppression and fragment merging, introducing sensitivity to post-processing. Block-level evaluation also measures local predictions rather than complete buildings. As Fig.~\ref{fig:teaser} illustrates, practical urban processing instead requires each building to be predicted and assessed as a complete object, motivating structure-aware training regions and unified scene-level evaluation.

Beyond separating individual buildings, recognizing their functions remains underexplored in urban instance segmentation \cite{Yang2023UrbanBIS}. Functional classification faces geometric and appearance ambiguity alongside long-tailed categories: similar footprints or facades can represent different uses, while the same category varies across neighborhoods. Acquisition density, viewpoint, and reconstruction quality further complicate recognition. Surrounding structures and spatial relationships can complement ambiguous building-body evidence. Combining local context, class-balanced supervision, and discriminative representation learning may therefore improve recognition of underrepresented and easily confused functions.

To overcome partition-induced fragmentation and functional ambiguity, we propose a framework coupling structure-preserving instance segmentation with contextual function recognition (Fig.~\ref{fig:pipeline}). Our adaptive region-dividing strategy transfers BEV semantic cues to 3D, constructing building-aligned training regions while retaining unassigned points for scene coverage. Training is thus guided by building structure, whereas complete scenes are directly processed and evaluated without block merging. Each segmented building is then interpreted jointly with its surroundings. To address long-tailed categories and ambiguous functions, we introduce spatially-supervised contrastive learning alongside class-balanced supervision. The contrastive objective gives greater weight to spatially proximate, same-category buildings within a scene, encouraging compact functional representations and separation between easily confused categories.

The contributions are summarized as follows.
\begin{itemize}
    \item We propose a semantics-guided adaptive region dividing strategy for structure-aligned training and a scene-level protocol evaluating complete building instances in their original scene space.
    \item We put forward a novel contextual fine-grained classifier combining class-balanced supervision and spatially-supervised contrastive learning to improve building-function representations.
    \item Extensive experiments on UrbanBIS and STPLS3D demonstrate our method's advantages over existing SOTA methods in scene-level instance segmentation and fine-grained function classification.
\end{itemize}

\section{Related Work}

\textbf{Urban Scene Point Cloud Understanding}.
Urban benchmarks support semantic and instance understanding of complex outdoor scenes. STPLS3D provides real and synthetic aerial photogrammetry point clouds with semantic and instance annotations \cite{Chen2022STPLS3D}, while UrbanBIS focuses on large-scale building instances and fine-grained categories \cite{Yang2023UrbanBIS}. Multimedia research explores self-supervised intra-modal and cross-modal contrastive representations for point-cloud understanding \cite{Wu2024IntraCrossModal}. From dense residential areas to campuses, variations in building scale, spacing, and surrounding context complicate region organization. Spatial grouping has also been studied through supervised contrastive learning for indoor oversegmentation \cite{Sun2025Oversegmentation}. Urban pipelines commonly partition scenes; B-Seg uses building-oriented regions to reduce fragmentation \cite{Yang2023UrbanBIS}, but predefined blocks can still yield partial instances and partition-dependent metrics. Vision foundation models offer semantic regions: Grounding DINO localizes open-vocabulary objects, while SAM and SAM 2 predict prompted masks \cite{liu2023grounding,kirillov2023segany,Ravi2024SAM2,ren2024grounded}. Remote-sensing adaptations improve prompting, multiscale features, and automated annotation for 2D segmentation \cite{Chen2024RSPrompter,Zhou2024MeSAM,Wang2023SAMRS}. We use grounded 2D perception only to construct structure-aligned 3D training regions. Final instances are predicted by the 3D network and evaluated in the original scene, with contextual representations supporting functional interpretation beyond geometric separation.

\textbf{3D Point Cloud Instance Segmentation}.
Existing methods follow detection-, grouping-, kernel-, or query-based paradigms. Detection and grouping construct proposals or cluster points through learned offsets and affinities \cite{Yang2019BoNet,Jiang2020PointGroup}; ISBNet decodes masks with instance-aware dynamic kernels \cite{Ngo2023ISBNet}. Query-based frameworks formulate segmentation as mask classification \cite{Carion2020DETR,Cheng2022Mask2Former}, as in Mask3D, OneFormer3D, and Relation3D \cite{Schult2023Mask3D,Kolodiazhnyi2024OneFormer3D,Lu2025Relation3D}. Learnable queries interact with scene features and directly predict masks. MAFT, Spherical Mask, EASE, and IKNE further explore attention-free decoding, spherical representations, semantic boundary cues, and discriminative instance features \cite{lai2023maft,shin2024spherical,roh2024EASE,roh2025IKNE}. Many methods were developed on bounded indoor benchmarks such as ScanNet \cite{Dai2017ScanNet}. These advances primarily concern instance representation and mask decoding. Urban scenes introduce larger objects, irregular coverage, and more points, often requiring partitions that fragment buildings and create duplicate predictions. Our work therefore focuses on structure-aware training regions and direct scene-level evaluation of complete buildings.

\textbf{Point Cloud Fine-grained Classification}.
Fine-grained recognition distinguishes similar subcategories using subtle cues. In image recognition, bilinear interactions and supervised contrastive learning improve feature modeling and inter-class separation \cite{Lin2015BilinearCNN,Khosla2020SupervisedContrastive}. Point-cloud recognition additionally faces sparsity, irregular sampling, and density variation. PointNet-style, graph-based, and transformer encoders address irregular geometry through aggregation, neighborhood modeling, and long-range interactions \cite{Qi2017PointNet,Qi2017PointNetPlusPlus,Wang2019DynamicGraphCNN,Zhao2021PointTransformer,Wu2024PointTransformerV3}; ScanObjectNN highlights clutter and background context in real-world recognition \cite{Uy2019ScanObjectNN}. However, these recognition studies largely focus on objects or indoor settings. Building-function studies commonly use aerial or street-view imagery and geospatial information \cite{Zhou2023BuildingUse,Kang2018BuildingInstanceClassification,Wang2021BuildingFunctionMapping}. UrbanBIS supports point-cloud-based functional classification \cite{Yang2023UrbanBIS}, but similar external forms and long-tailed categories complicate recognition. Nearby roads, facilities, and building arrangements can provide complementary functional cues. Class-balanced objectives increase the contribution of underrepresented categories \cite{Cui2019ClassBalanced}, while supervised contrastive learning attracts same-category instances and separates different categories \cite{Khosla2020SupervisedContrastive}. Such reweighting changes category contributions without altering the inference architecture. Standard supervised contrastive learning treats same-category positives equally. We instead use spatial proximity as a training-only weighting signal for same-category, same-scene pairs that may share functional environments.

\section{Method}

Our framework (Fig.~\ref{fig:pipeline}) performs building instance segmentation and fine-grained classification on an urban point cloud $P \in \mathbb{R}^{N \times (3+F)}$, with $N$ points and $F$ additional attributes. Segmentation trains on manageable, structure-aligned building-aware and unassigned blocks but directly processes unpartitioned $P$ for complete-instance evaluation without block merging. Each predicted building is combined with context for function classification.

\subsection{Adaptive Region Dividing}

To reduce building truncation, BEV-guided adaptive region dividing constructs building-aware blocks from detected regions and unassigned blocks from the remaining points. Together, they preserve building structure and scene coverage at controllable input sizes.

\subsubsection{BEV-based Building Candidate Perception}
We project $P=\{p_i\}_{i=1}^{N}$, with $p_i \in \mathbb{R}^{3+F}$, onto the bird's-eye-view plane to obtain the RGB image $I_{\mathrm{BEV}}$, retaining point-to-pixel correspondences for back-projection. Pixel colors average the RGB values of their mapped points, preserving scene structure and color information.

Within the Grounded SAM framework, Grounding DINO followed by SAM 2 perceives building regions in $I_{\mathrm{BEV}}$ \cite{kirillov2023segany,liu2023grounding,ren2024grounded,Ravi2024SAM2}. Given the prompt $t=\text{``building''}$, Grounding DINO predicts boxes $b_k$ with confidence scores $\gamma_k$, and SAM 2 constructs core masks $C_k$ from these spatial prompts. The complete set of BEV building masks is denoted by $\mathcal C^{\mathrm{BEV}}=\{C_k\}_{k=1}^{K}$, where $K$ is the number of retained building candidates. Each candidate is represented as $R_k=(C_k,E_k)$, where $E_k$ expands $b_k$ by a predefined metric distance to include context.

\subsubsection{Building-aware Block Construction}
Using the retained pixel correspondence, points are mapped from the point cloud to each candidate $R_k$. Because candidate regions may overlap in BEV space, a block cannot be defined by simple region membership. We therefore define a point-level assignment function $a_i$: $a_i=k$ assigns $p_i$ to candidate $k$, whereas $a_i=-1$ denotes an unassigned point. Core region $C_k$ first determines building-body points, after which $E_k$ supplies local context. For a point covered by multiple core masks, the candidate maximizing $\gamma_k+0.1\eta_k$ is selected, where $\eta_k$ is the SAM-2 mask score. Expanded regions process only points not assigned by any core mask, and overlapping context points are assigned to the candidate with the nearest horizontal centroid. The resulting block $B_k$ better follows building structure than a fixed window but need not correspond one-to-one with an instance due to missed, merged, or split 2D detections.

\subsubsection{Unassigned Block Construction}
To retain missed building regions and non-building context for learning the complete data distribution and building boundaries, all points with $a_i=-1$ are partitioned into non-overlapping regular BEV windows, forming $B_{\mathrm{unassigned}}$.

The complete training set is $B_{\mathrm{all}}=B_{\mathrm{build}}\cup B_{\mathrm{unassigned}}$, where $B_{\mathrm{build}}$ contains the building-aware blocks and $B_{\mathrm{unassigned}}$ retains the remaining scene coverage.

\subsection{Scene-level Building Instance Segmentation and Evaluation}

The model trains on adaptive blocks but evaluates $P$ directly under the scene-level protocol without merging.

\subsubsection{Sparse Voxel Encoding and Query-based Prediction}
For the $n$-th training block $B_n\in B_{\mathrm{all}}$, voxelization $\mathcal V(\cdot)$ produces sparse tokens $V_n=\mathcal V(B_n)$. A sparse convolutional backbone encodes $V_n$ into fused features $Z_n^{s}$ for decoding a fixed number $Q$ of instance queries. Each outputs building/no-object logits, sparse-token mask logits, and a separate mask-quality score. Inverse mapping restores mask probabilities to original points, yielding $\hat{I}^{B_n}=\{(\hat{I}_{n,q},s_{n,q})\}_{q=1}^{Q}$, where $\hat{I}_{n,q}$ is the $q$-th building mask on $B_n$ and $s_{n,q}$ its quality score.

\subsubsection{Query Matching and Instance Training Objective}
For a block $B_n$ with $N_n$ points, query $q$ predicts building/no-object probabilities $\hat{\mathbf p}_{n,q}$, point-mask probabilities $\hat{\mathbf m}_{n,q}\in[0,1]^{N_n}$, and mask-quality score $\hat s_{n,q}$. Let $\mathcal{G}_n=\{(\mathbf{m}^{*}_{n,j},y^{*}_{n,j})\}_{j=1}^{M_n}$ contain its $M_n$ ground-truth masks and building labels. Following DETR-style set prediction and mask-classification objectives \cite{Carion2020DETR,Cheng2022Mask2Former,Schult2023Mask3D}, Hungarian matching \cite{Kuhn1955HungarianMethod} establishes a one-to-one assignment using classification, binary cross-entropy (BCE), and Dice costs.
After matching, matched query $q$ receives label $y^{\pi}_{n,q}$ and ground-truth mask $\mathbf m^{\pi}_{n,q}$; unmatched queries receive no-object. Let $\mathcal M_n$ contain matched queries. The class weight $\alpha_y$ is $1.0$ for buildings and $0.05$ for no-object. Following IoU-guided proposal scoring in point-cloud instance segmentation \cite{Jiang2020PointGroup,Vu2022SoftGroup}, the detached target $s^{*}_{n,q}$ is the point-level IoU between $\mathbf m^{\pi}_{n,q}$ and $\hat{\mathbf m}_{n,q}$ thresholded at $0.5$, and $\mathcal S_n$ contains matches with $s^{*}_{n,q}>0.5$. The four adopted losses are
\begin{equation}
\begin{aligned}
\mathcal L_{\mathrm{cls}}&=-\frac{1}{Q}\sum_{q=1}^{Q}
\alpha_{y^{\pi}_{n,q}}\log\hat p_{n,q,y^{\pi}_{n,q}},\\
\mathcal L_{\mathrm{bce}}&=\frac{1}{|\mathcal M_n|}\sum_{q\in\mathcal M_n}
\operatorname{BCE}(\hat{\mathbf m}_{n,q},\mathbf m^{\pi}_{n,q}),\\
\mathcal L_{\mathrm{dice}}&=\frac{1}{|\mathcal M_n|}\sum_{q\in\mathcal M_n}
\operatorname{Dice}(\hat{\mathbf m}_{n,q},\mathbf m^{\pi}_{n,q}),\\
\mathcal L_{\mathrm{score}}&=\frac{1}{|\mathcal S_n|}\sum_{q\in\mathcal S_n}
(\hat s_{n,q}-s^{*}_{n,q})^2.
\end{aligned}
\label{eq:instance_loss_terms}
\end{equation}
$\mathcal L_{\mathrm{score}}$ is zero when $\mathcal S_n$ is empty. The classification, BCE, and Dice terms follow standard query-based mask supervision \cite{Cheng2022Mask2Former,Schult2023Mask3D}, while the score term calibrates predicted mask quality. After batch averaging, the final instance training objective is
\begin{equation}
\mathcal L_{\mathrm{ins}}=
0.5\mathcal L_{\mathrm{cls}}+
\mathcal L_{\mathrm{bce}}+
\mathcal L_{\mathrm{dice}}+
0.5\mathcal L_{\mathrm{score}}.
\label{eq:instance_training_objective}
\end{equation}

\subsubsection{Scene-level Inference and Evaluation}
Point cloud $P$ is voxelized and processed without BEV projection or adaptive dividing during evaluation. Sparse feature extraction, query decoding, and inverse mapping produce $\hat{I}^{P}=\{(\hat{I}^{P}_{q},s^{P}_{q})\}_{q=1}^{Q}$, where $\hat{I}^{P}_{q}$ denotes the $q$-th building mask on $P$, and $s^{P}_{q}$ is its mask-quality score. Query classification, mask-quality filtering, and thresholding yield scene-level predictions on $P$.

Predicted and ground-truth instances are matched in the same original point space using point-level $\mathrm{IoU}(\hat{I}^{P}_{q},I^{*,P}_{j})=|\hat{I}^{P}_{q}\cap I^{*,P}_{j}|/|\hat{I}^{P}_{q}\cup I^{*,P}_{j}|$.
A prediction is a true positive if its IoU with an unmatched ground-truth instance exceeds threshold $\theta$. Following common instance segmentation protocols, we report AP, AP$_{50}$, and AP$_{25}$. Unlike block-level evaluation, scene-level prediction and metric computation operate on complete instances in $P$, avoiding the effects of block truncation, duplicate predictions, and cross-block merging strategies. As a transferability analysis, we apply adaptive region dividing to other instance segmentation models and evaluate them under the same scene-level protocol.

\subsection{Fine-grained Building Function Classification}

Fine-grained classification learns a building-level representation from each instance. Training uses ground-truth instances to avoid segmentation noise, whereas inference uses instances predicted by the preceding segmentation module.

\subsubsection{Core-context Input and Feature Encoding}
Expanding the horizontal bounding box of instance $I_j$ by a fixed distance yields context $\mathcal{N}_j$ and input $X_j=I_j\cup\mathcal{N}_j$. Nearby roads, facilities, and structures can provide functional cues beyond building geometry. Indicator $r_{j,i}=1$ marks target-building core points and $r_{j,i}=0$ marks non-core context, distinguishing their roles for the encoder. Without context, $\mathcal{N}_j=\varnothing$ and all indicators are one. Each point in $X_j$ uses local coordinates, RGB, and the binary core-context indicator $r_{j,i}$.

We adopt Point Transformer V3 as the point-cloud encoder \cite{Wu2024PointTransformerV3}. Local coordinates model building geometry and local neighborhoods, whereas global scene coordinates are excluded from the semantic input to avoid direct location-dependent classification. Following the PTv3 classification design, the encoder extracts point features $\mathbf{u}_{j,i}$ from $X_j$, which are aggregated by concatenated max and mean pooling to obtain the building representation $\mathbf{h}_j$.

\subsubsection{Class-Balanced Weighted Cross-Entropy}
A classification head maps $\mathbf{h}_j$ to logits over $K_{\mathrm{cls}}$ categories. We optimize weighted cross-entropy $\mathcal{L}_{\mathrm{wce}}$ to reduce frequent-category dominance through loss reweighting \cite{Cui2019ClassBalanced}. For category $c$ with $n_c$ training instances, we use the unnormalized weight $\tilde{w}_c=\min\!\left(3.0,[\log(1.2+n_c)]^{-1}\right)$, followed by mean normalization $w_c=\tilde{w}_c/(K_{\mathrm{cls}}^{-1}\sum_{r=1}^{K_{\mathrm{cls}}}\tilde{w}_r)$. This increases the relative contribution of underrepresented categories while maintaining a stable loss scale.

\subsubsection{Spatially-Supervised Contrastive Loss}
In parallel with the classification head, the projection head $g_{\mathrm{proj}}(\cdot)$ maps $\mathbf{h}_j$ to the $\ell_2$-normalized embedding $\mathbf{z}_j=g_{\mathrm{proj}}(\mathbf{h}_j)/\|g_{\mathrm{proj}}(\mathbf{h}_j)\|_2$.

Category supervision does not organize the building-level embedding space, so we introduce a spatially-supervised contrastive loss. Same-category buildings within a scene form positive pairs because they may share local functional context, while different-category instances serve as negatives. Unlike standard supervised contrastive learning, spatially closer positive pairs receive larger weights.

For a same-category positive pair $(j,u)$, let $\mathbf c_j$ and $\mathbf c_u$ denote horizontal centers and $\sigma$ the spatial decay scale. Its spatial weight $\omega_{j,u}$ is
\begin{equation}
\omega_{j,u}
=
\exp
\left(
-
\frac{
\|\mathbf{c}_j-\mathbf{c}_u\|_2^2
}{
\sigma^2
}
\right).
\end{equation}
Because the embeddings are $\ell_2$-normalized, their similarity is $\mathrm{sim}(\mathbf{z}_j,\mathbf{z}_u)=\mathbf{z}_j^{\top}\mathbf{z}_u$.

Let $\mathrm{Pos}(j)$ contain same-category instances from the same scene, $\tau$ be the contrastive temperature, and $\epsilon$ ensure numerical stability; the denominator spans all non-anchor batch instances. The exponential similarity $\kappa_{j,v}$, pairwise contrastive term $\ell_{j,u}$, and spatial normalizer $Z_j$ are
\begin{equation}
\begin{aligned}
\kappa_{j,v}
&=
\exp(\mathrm{sim}(\mathbf{z}_j,\mathbf{z}_v)/\tau),\\
\ell_{j,u}
&=
-\log
\frac{\kappa_{j,u}}{\sum_{v \neq j}\kappa_{j,v}},\\
Z_j
&=
\sum_{u \in \mathrm{Pos}(j)}\omega_{j,u}+\epsilon.
\end{aligned}
\end{equation}
Let $A$ contain anchors with valid positives; other anchors are skipped. The spatially-supervised contrastive loss $\mathcal L_{\mathrm{sp}}$ is
\begin{equation}
\mathcal{L}_{\mathrm{sp}}
=
\frac{1}{|A|}
\sum_{j \in A}
\frac{1}{Z_j}
\sum_{u \in \mathrm{Pos}(j)}
\omega_{j,u}\ell_{j,u}.
\end{equation}
The final objective is $\mathcal{L}=\mathcal{L}_{\mathrm{wce}}+\lambda_{\mathrm{sp}}\mathcal{L}_{\mathrm{sp}}$.
Scene-aware class-balanced sampling selects categories and multiple instances from one scene to provide valid positive pairs while reducing class imbalance. Global coordinates are used to extract local context at training and inference and to compute spatial weights only during training. After context extraction, coordinates are recentered; the classifier receives neither absolute locations nor spatial-neighbor relationships.

\section{Experiments}

\subsection{Experiment Settings}

\textbf{Datasets.}
We conduct experiments on two large-scale urban point cloud datasets, UrbanBIS and STPLS3D.

\textit{UrbanBIS} \cite{Yang2023UrbanBIS} provides real-world aerial photogrammetry point clouds with building instances and eight functional categories: Commercial (Co), Residential (Re), Office (Of), Cultural (Cu), Transportation (Tr), Municipal (Mu), Temporary (Te), and Unclassified (Un). Same-scene experiments use Qingdao, Wuhu, Longhua, and Campus; Campus merges Yuehai and Lihu while preserving their original split membership. Their spatially disjoint train/validation/test scene-unit counts are 20/6/10, 26/13/13, 24/12/12, and 36/8/12, respectively. Models train within each scene and are evaluated on complete test scene units. Yingrenshi has one annotated test scene unit, used only as the held-out cross-scene target.

\textit{STPLS3D} \cite{Chen2022STPLS3D} is a large-scale aerial photogrammetry point cloud dataset containing synthetic and real-world urban regions with semantic and instance annotations. We use the official training split for training and report results on the validation split by directly processing each complete scene. Only building instances are retained as supervision and prediction targets, while all non-building classes are ignored. Since STPLS3D does not provide fine-grained building function labels, it is used only for building instance segmentation.

\textbf{Comparison Methods.}
For building instance segmentation, we compare against representative existing methods listed in the corresponding result tables. Unless otherwise stated, all baseline results are obtained from our re-implementations or adapted official implementations under the same data splits and evaluation protocols; they are not copied from the originally reported benchmark values. Our model directly processes each original test scene, whereas methods restricted to block-level inference first predict instances within their test blocks. To evaluate all methods in the same scene-level space, we project their block masks to the original point-index space. Because neighboring blocks may overlap and predict the same building multiple times, duplicate candidates are merged after projection. Candidate instances are sorted by confidence and compared using point-level IoU and the containment ratio $\mathrm{Contain}(I_a,I_b)=|I_a\cap I_b|/\min(|I_a|,|I_b|)$.
Candidates with scores below $0.05$ are removed. Two candidates are merged when their point-level IoU is at least $0.50$ or their containment ratio is at least $0.60$; their point-index sets are combined by union and the higher score is retained. Candidates with at most 8,000 points are handled by the fragment-attachment step, and final masks with fewer than 100 points are discarded. We use these fixed settings for all block-based methods and scenes without method-specific tuning. This recovery is applied only when a comparison method cannot natively process an original scene; our method requires no merging.

For fine-grained classification, we compare our complete PTv3-based classifier against representative point-cloud classification baselines. All methods use the same UrbanBIS data splits, category mapping, input point count, and evaluation metrics. In the main comparison, every classifier is trained on ground-truth building instances and evaluated on the same instances predicted by the preceding segmentation module. The comparison methods follow their standard classification configurations, whereas Ours (PTv3) additionally incorporates core-context input, class-balanced weighted cross-entropy, and spatially-supervised contrastive loss. This is a method-level comparison, not a backbone ablation.

\textbf{Implementation Details.}
Adaptive region dividing uses a BEV resolution of $0.5$ m/pixel and expands each detected building box by 10 m. Building candidates are extracted using the Grounding DINO Swin-T OGC checkpoint (\texttt{groundingdino\_swint\_ogc.pth}) and the SAM 2.1 Hiera-L checkpoint (\texttt{sam2.1\_hiera\_large.pt}); the box and text confidence thresholds are both set to $0.25$. Box NMS and mask deduplication use IoU thresholds of $0.35$ and $0.75$, respectively. Unassigned points in UrbanBIS are partitioned by non-overlapping $100\,\mathrm{m}\times100\,\mathrm{m}$ windows. For UrbanBIS instance segmentation, the voxel size is $0.2$ m, 400 instance queries are decoded, and at most 200,000 points are sampled per training input. Across all UrbanBIS test scene units, the maximum number of ground-truth building instances in a single input is 75, well below the query number, confirming that the query budget does not truncate the ground-truth instance set. We train for 20,000 iterations with AdamW, a learning rate of $10^{-4}$, weight decay $0.05$, and a batch size of one per GPU on two NVIDIA Quadro P6000 GPUs using mixed precision. For all UrbanBIS scenes, we evaluate the final 20,000-iteration checkpoints without validation-based checkpoint selection. For STPLS3D, we evaluate the final 20,000-iteration checkpoint with a voxel size of $0.333$ m and 160 queries.

For fine-grained classification, each input contains 2,048 points with a 10 m context expansion and is encoded by Point Transformer V3 \cite{Wu2024PointTransformerV3}. Training uses AdamW with a learning rate of $10^{-4}$ for at most 200 epochs and selects the checkpoint by validation Macro-F1. Unless otherwise specified, we use random seed 42. We set the contrastive temperature $\tau$ to $0.1$, the spatial scale $\sigma$ to $30\,\mathrm{m}$, and the contrastive-loss weight $\lambda_{\mathrm{sp}}$ to $0.1$.

\textbf{Metrics.}
For building \textit{instance segmentation}, we report AP averaged over IoU thresholds $0.50$--$0.90$ in steps of $0.05$, and AP$_{50}$ and AP$_{25}$ at thresholds $0.50$ and $0.25$, respectively. All UrbanBIS main, adaptive-region, and cross-scene comparisons, together with STPLS3D results, use scene-level evaluation in the original target-scene space.
For building \textit{fine-grained classification}, we report overall accuracy (OA), mean class accuracy (MCA), and Macro-F1. Since building function categories in UrbanBIS are highly imbalanced, OA alone cannot fully reflect classification performance. MCA and Macro-F1 are therefore used to evaluate the behavior of the classifier on long-tailed and easily confused categories. For evaluation on predicted instances, we compute point-level IoU between predicted and ground-truth instances and establish one-to-one correspondences by maximum-IoU matching; if multiple predictions correspond to the same ground-truth instance, only the prediction with the highest IoU is retained. Only pairs with IoU $\geq 0.50$ are accepted, and OA, MCA, and Macro-F1 are computed over these matched predicted instances. The resulting ground-truth instance coverage is $85.25\%$, indicating the proportion of ground-truth buildings included in the classification evaluation.

\subsection{Experiment Results}

\subsubsection{Scene-level Instance Segmentation}

\begin{table*}[!t]
\centering
\caption{Scene-Level Building Instance Segmentation Results on UrbanBIS. All Methods Are Evaluated on the Original Scenes.}
\label{tab:urbanbis_full_scene_seg}
\setlength{\tabcolsep}{6pt}
\begin{tabular}{c|ccc|ccc|ccc|ccc}
\toprule
\multirow{2}{*}{Method}
& \multicolumn{3}{c|}{Qingdao}
& \multicolumn{3}{c|}{Wuhu}
& \multicolumn{3}{c|}{Longhua}
& \multicolumn{3}{c}{Campus} \\
\cmidrule(lr){2-4}
\cmidrule(lr){5-7}
\cmidrule(lr){8-10}
\cmidrule(lr){11-13}
& AP & AP$_{50}$ & AP$_{25}$
& AP & AP$_{50}$ & AP$_{25}$
& AP & AP$_{50}$ & AP$_{25}$
& AP & AP$_{50}$ & AP$_{25}$ \\
\midrule
PointGroup\cite{Jiang2020PointGroup} & 0.256 & 0.310 & 0.432 & 0.436 & 0.532 & 0.616 & 0.203 & 0.299 & 0.384 & 0.263 & 0.397 & 0.485 \\
HAIS\cite{Chen2021HAIS} & 0.153 & 0.235 & 0.290 & 0.164 & 0.258 & 0.330 & 0.066 & 0.153 & 0.226 & 0.164 & 0.208 & 0.299 \\
SoftGroup\cite{Vu2022SoftGroup} & 0.161 & 0.228 & 0.304 & 0.158 & 0.267 & 0.341 & 0.058 & 0.147 & 0.219 & 0.171 & 0.214 & 0.288 \\
DyCo3D\cite{He2021dyco3d} & 0.146 & 0.242 & 0.281 & 0.169 & 0.251 & 0.322 & 0.071 & 0.159 & 0.231 & 0.157 & 0.203 & 0.307 \\
DKNet\cite{wu2022dknet} & 0.155 & 0.237 & 0.295 & 0.162 & 0.263 & 0.335 & 0.063 & 0.150 & 0.222 & 0.166 & 0.219 & 0.294 \\
B-Seg\cite{Yang2023UrbanBIS} & 0.402 & 0.492 & 0.558 & 0.471 & 0.595 & 0.650 & 0.223 & 0.314 & 0.440 & 0.392 & 0.474 & 0.553 \\
Mask3D\cite{Schult2023Mask3D} & 0.421 & 0.509 & 0.575 & 0.486 & 0.611 & 0.669 & 0.238 & 0.329 & 0.454 & 0.427 & 0.491 & 0.566 \\
OneFormer3D\cite{Kolodiazhnyi2024OneFormer3D} & 0.414 & 0.501 & 0.563 & 0.478 & 0.604 & 0.658 & 0.229 & 0.322 & 0.447 & \underline{0.431} & 0.482 & 0.559 \\
MAFT\cite{lai2023maft} & 0.427 & \underline{0.516} & \underline{0.581} & \underline{0.491} & 0.617 & 0.674 & 0.242 & \underline{0.335} & \underline{0.458} & 0.412 & \underline{0.496} & \underline{0.571} \\
Relation3D\cite{Lu2025Relation3D} & \underline{0.448} & 0.507 & 0.572 & 0.483 & \underline{0.638} & \underline{0.684} & \underline{0.251} & 0.326 & 0.451 & 0.405 & 0.487 & 0.564 \\
Ours & \textbf{0.558} & \textbf{0.687} & \textbf{0.714} & \textbf{0.589} & \textbf{0.663} & \textbf{0.747} & \textbf{0.305} & \textbf{0.422} & \textbf{0.536} & \textbf{0.453} & \textbf{0.511} & \textbf{0.594} \\
\bottomrule
\end{tabular}
\vspace{-0.3cm}
\end{table*}

\begin{figure*}[!t]
    \centering
    \includegraphics[width=0.95\textwidth]{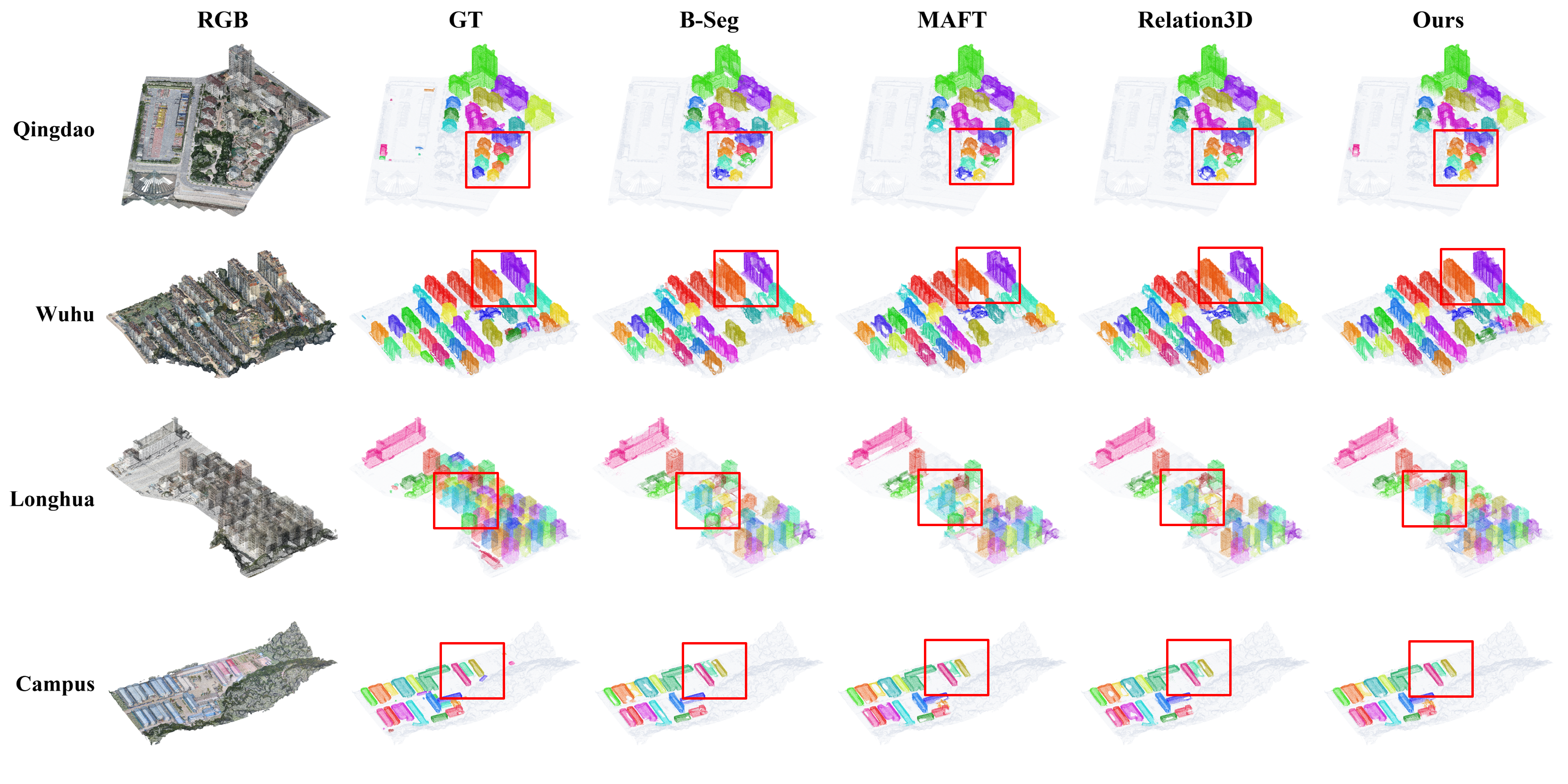}
    \vspace{-0.3cm}
    \caption{Qualitative comparison of scene-level building instance segmentation results on UrbanBIS. The rows correspond to the Qingdao, Wuhu, Longhua, and Campus scenes, while the columns show the RGB point clouds, ground truth (GT), and predictions from B-Seg, MAFT, Relation3D, and our method. Different colors denote individual building instances against the muted scene background, and the red boxes highlight representative regions for comparing instance completeness and the separation of adjacent buildings.}
    \label{fig:urbanbis_results}
    \vspace{-0.3cm}
\end{figure*}

\textbf{Scene-level Results on UrbanBIS.}
\Cref{tab:urbanbis_full_scene_seg} compares methods on Qingdao, Wuhu, Longhua, and Campus. Our model trains on adaptive blocks and tests on complete scenes; block-based baselines use the recovery procedure specified above. All results measure complete instances in the original scene space.
Our method achieves the best performance on all four scenes under this scene-level protocol (\cref{tab:urbanbis_full_scene_seg}), which evaluates complete buildings including those crossing conventional block boundaries.
In \cref{fig:urbanbis_results}, highlighted predictions follow ground-truth building extents more closely and better separate adjacent instances across high-rise, residential, and campus layouts.

\begin{table}[!t]
\centering
\caption{Scene-Level Building Instance Segmentation Results on the STPLS3D Validation Set.}
\label{tab:stpls3d_seg}
\begin{tabular}{cccc}
\toprule
Method & AP & AP$_{50}$ & AP$_{25}$ \\
\midrule
SoftGroup\cite{Vu2022SoftGroup} & 0.435 & 0.609 & 0.693 \\
ISBNet\cite{Ngo2023ISBNet} & 0.467 & 0.612 & 0.741 \\
Mask3D\cite{Schult2023Mask3D} & 0.753 & \underline{0.863} & 0.902 \\
Spherical Mask\cite{shin2024spherical} & 0.512 & 0.664 & 0.770 \\
TD3D\cite{td3d} & 0.518 & 0.659 & 0.775 \\
OneFormer3D\cite{Kolodiazhnyi2024OneFormer3D} & 0.570 & 0.837 & 0.883 \\
EASE\cite{roh2024EASE} & \underline{0.755} & 0.848 & \underline{0.910} \\
IKNE\cite{roh2025IKNE} & 0.751 & \textbf{0.869} & 0.909 \\
Ours & \textbf{0.769} & 0.854 & \textbf{0.932} \\
\bottomrule
\end{tabular}
\vspace{-0.5cm}
\end{table}

\begin{figure*}[!t]
    \centering
    \includegraphics[width=0.95\textwidth]{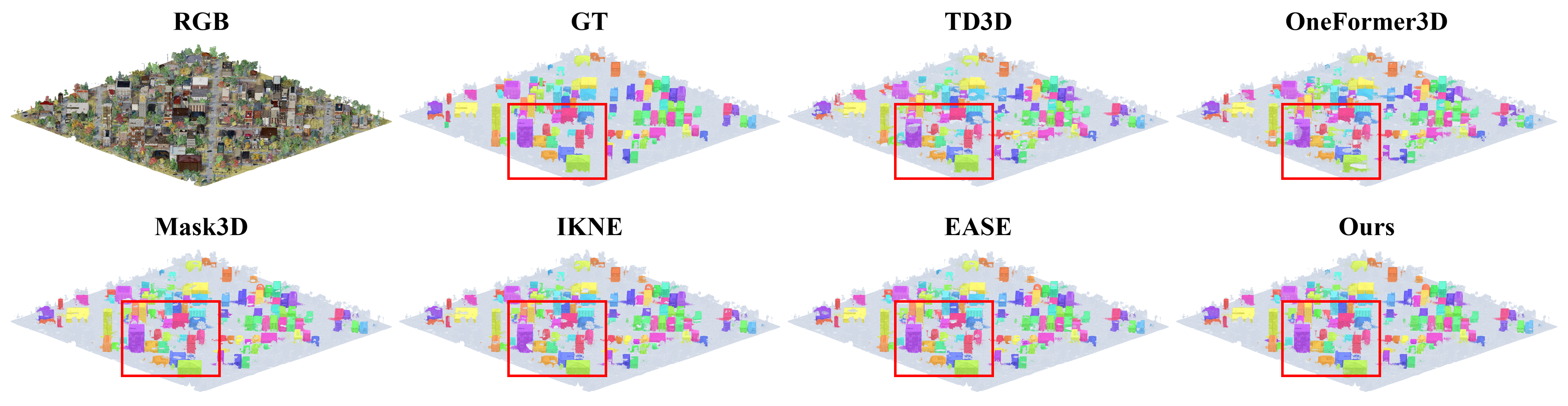}
    \vspace{-0.3cm}
    \caption{Qualitative comparison of scene-level building instance segmentation results on Area 20 of the STPLS3D validation set. The panels show the RGB point cloud, ground truth (GT), and predictions from TD3D, OneFormer3D, Mask3D, IKNE, EASE, and our method. Different colors denote individual building instances against the muted scene background, and the red boxes mark the same representative region for detailed comparison.}
    \vspace{-0.3cm}
    \label{fig:stpls3d_results}
\end{figure*}

\textbf{Scene-level Results on STPLS3D.}
\Cref{tab:stpls3d_seg} reports scene-level building segmentation on the STPLS3D validation set, ignoring non-building classes and invalid instance labels during training and evaluation.
STPLS3D complements UrbanBIS with a different urban point-cloud distribution. Our method achieves the highest AP and AP$_{25}$, supporting its effectiveness beyond UrbanBIS.
For Area 20 (\cref{fig:stpls3d_results}), the highlighted dense region shows better-preserved building extents and boundaries, with fewer merged or fragmented predictions.

\subsubsection{Fine-grained Classification}
\textbf{Fine-grained Results on UrbanBIS.}
\Cref{tab:fine_grained_main} compares per-class accuracy and aggregate metrics. Our method achieves the strongest overall performance and remains leading or competitive across most functions. Its advantage in OA, MCA, and Macro-F1 indicates more balanced recognition under the long-tailed distribution, rather than gains confined to frequent categories.

\begin{figure}[!t]
    \centering
    \includegraphics[width=\columnwidth]{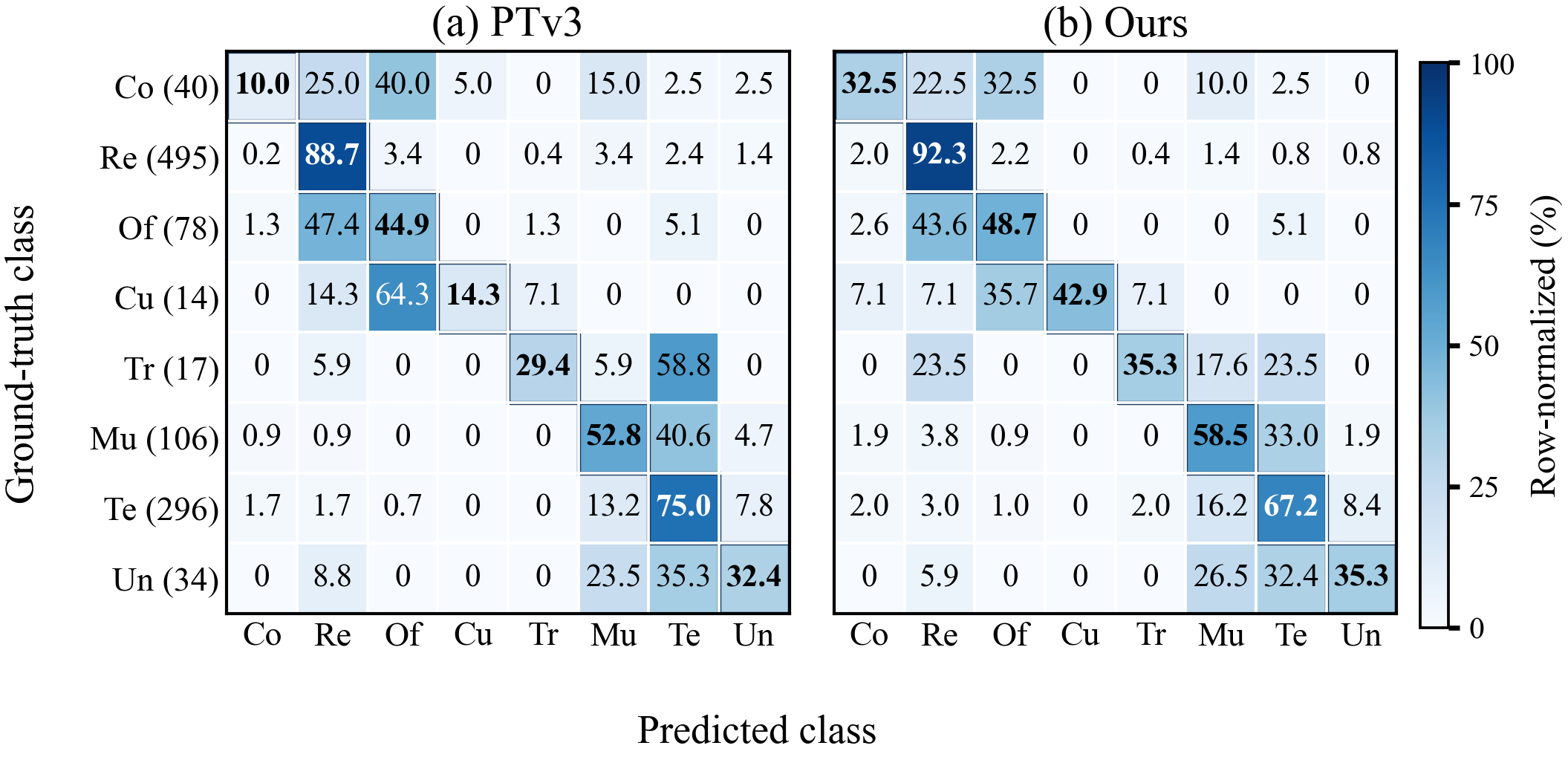}
    \vspace{-0.6cm}
    \caption{Row-normalized UrbanBIS building-function confusion matrices for (a) the PTv3 baseline and (b) our method. Rows and columns denote ground-truth and predicted categories, respectively; each row sums to 100\%, class sizes are shown in parentheses, and abbreviations are defined in Table~\ref{tab:fine_grained_main}.}
    \label{fig:confusion_matrix}
    \vspace{-0.3cm}
\end{figure}

Compared with PTv3, the confusion matrices (\cref{fig:confusion_matrix}) show improved diagonal accuracy for seven categories, particularly Commercial and Cultural buildings previously confused with Office. Misclassifications of Transportation and Municipal buildings as Temporary also decrease substantially. Together with the component ablations, these improvements support combining contextual cues, class-balanced supervision, and spatially-supervised contrastive learning to distinguish ambiguous functions. Temporary is the only category with lower diagonal accuracy, and its overlap with Municipal and Unclassified remains a challenge.

\begin{table*}[!t]
\centering
\caption{Fine-grained classification comparison on UrbanBIS. All methods are trained on ground-truth instances and evaluated on the same first-stage predicted instances.}
\label{tab:fine_grained_main}
\setlength{\tabcolsep}{3pt}
\begin{tabular}{ccccccccc|ccc}
\toprule
Method & Co & Re & Of & Cu & Tr & Mu & Te & Un & OA & MCA & Macro-F1 \\
\midrule
PointNet++\cite{Qi2017PointNetPlusPlus} & \underline{0.275} & 0.709 & 0.333 & 0.143 & 0.059 & 0.519 & 0.172 & 0.500 & 0.476 & 0.339 & 0.277 \\
DGCNN\cite{Wang2019DynamicGraphCNN} & 0.100 & 0.723 & \textbf{0.615} & 0.214 & \underline{0.294} & 0.491 & 0.223 & 0.471 & 0.511 & 0.391 & 0.313 \\
PointMLP\cite{Ma2022PointMLP} & 0.200 & 0.590 & \underline{0.564} & 0.357 & 0.059 & 0.132 & 0.230 & \textbf{0.735} & 0.423 & 0.358 & 0.258 \\
PointNeXt-S\cite{Qian2022PointNeXt} & 0.225 & 0.495 & \underline{0.564} & 0.214 & 0.059 & 0.359 & 0.196 & \underline{0.618} & 0.388 & 0.341 & 0.257 \\
PTv3\cite{Wu2024PointTransformerV3} & 0.100 & \underline{0.887} & 0.449 & 0.143 & \underline{0.294} & \underline{0.528} & \textbf{0.750} & 0.324 & \underline{0.717} & \underline{0.434} & \underline{0.448} \\
PointMamba\cite{Liang2024PointMamba} & \underline{0.275} & 0.647 & 0.410 & 0.286 & 0.118 & 0.198 & 0.287 & 0.382 & 0.452 & 0.325 & 0.276 \\
PointKAN\cite{Shi2025PointKAN} & 0.050 & 0.669 & 0.128 & \textbf{0.500} & 0.118 & 0.104 & 0.527 & 0.441 & 0.494 & 0.317 & 0.257 \\
Ours (PTv3) & \textbf{0.325} & \textbf{0.923} & 0.487 & \underline{0.429} & \textbf{0.353} & \textbf{0.585} & \underline{0.672} & 0.353 & \textbf{0.734} & \textbf{0.516} & \textbf{0.536} \\
\bottomrule
\end{tabular}
\vspace{-0.4cm}
\end{table*}

\subsection{Ablation Study}

\subsubsection{Scene-level Instance Segmentation}
\textbf{Ablation Study on Adaptive Region Dividing and Cross-Architecture Transferability.}
\Cref{tab:urbanbis_adaptive_seg} evaluates adaptive region dividing from two complementary perspectives. For the comparison models, we keep the network architecture, data splits, block-to-scene recovery procedure, and scene-level evaluation protocol unchanged, replacing only the original region-dividing scheme with adaptive region dividing. Because the region-dividing scheme determines both the training inputs and, for models restricted to partitioned inference, the test blocks, these comparisons quantify its overall contribution to the complete instance segmentation pipeline rather than its training-only effect. The consistent improvements obtained by different architectures under the same scene-level evaluation demonstrate the transferability and practical utility of the proposed strategy. For our model, both the Original and Adaptive settings directly process the same unpartitioned scenes at test time and differ only in their training partitions. This comparison therefore further isolates the training effect of adaptive blocks: they improve AP in all four scenes, with an average gain of approximately 0.023, indicating that structure-aligned regions facilitate learning more complete building representations.

\begin{table*}[!t]
\centering
\caption{Comparison of the Original Partitioning Schemes and the Proposed Adaptive Region-Dividing Strategy for Building Instance Segmentation on UrbanBIS under Scene-Level Evaluation.}
\label{tab:urbanbis_adaptive_seg}
\setlength{\tabcolsep}{2.6pt}
\begin{tabular}{cc|ccc|ccc|ccc|ccc}
\toprule
\multirow{2}{*}{Method}
& \multirow{2}{*}{Partition}
& \multicolumn{3}{c|}{Qingdao}
& \multicolumn{3}{c|}{Wuhu}
& \multicolumn{3}{c|}{Longhua}
& \multicolumn{3}{c}{Campus} \\
\cmidrule(lr){3-5}
\cmidrule(lr){6-8}
\cmidrule(lr){9-11}
\cmidrule(lr){12-14}
& & AP & AP$_{50}$ & AP$_{25}$
& AP & AP$_{50}$ & AP$_{25}$
& AP & AP$_{50}$ & AP$_{25}$
& AP & AP$_{50}$ & AP$_{25}$ \\
\midrule
\multirow{2}{*}{B-Seg\cite{Yang2023UrbanBIS}}
& Original & 0.402 & 0.492 & 0.558 & 0.471 & 0.595 & 0.650 & 0.223 & 0.314 & 0.440 & 0.392 & 0.474 & 0.553 \\
& Adaptive & 0.476 & 0.571 & 0.696 & 0.568 & 0.601 & 0.690 & 0.297 & 0.336 & 0.451 & 0.424 & 0.483 & 0.558 \\
\midrule
\multirow{2}{*}{Mask3D\cite{Schult2023Mask3D}}
& Original & 0.421 & 0.509 & 0.575 & 0.486 & 0.611 & 0.669 & 0.238 & 0.329 & 0.454 & 0.427 & 0.491 & 0.566 \\
& Adaptive & 0.486 & 0.589 & 0.707 & 0.584 & 0.623 & 0.701 & 0.253 & 0.348 & 0.467 & 0.439 & 0.508 & 0.577 \\
\midrule
\multirow{2}{*}{OneFormer3D\cite{Kolodiazhnyi2024OneFormer3D}}
& Original & 0.414 & 0.501 & 0.563 & 0.478 & 0.604 & 0.658 & 0.229 & 0.322 & 0.447 & 0.431 & 0.482 & 0.559 \\
& Adaptive & 0.480 & 0.580 & 0.699 & 0.577 & 0.628 & 0.724 & 0.295 & 0.339 & 0.469 & 0.440 & 0.491 & 0.570 \\
\midrule
\multirow{2}{*}{MAFT\cite{lai2023maft}}
& Original & 0.427 & 0.516 & 0.581 & 0.491 & 0.617 & 0.674 & 0.242 & 0.335 & 0.458 & 0.412 & 0.496 & 0.571 \\
& Adaptive & 0.489 & 0.576 & 0.701 & 0.581 & 0.654 & 0.717 & 0.289 & 0.342 & 0.483 & 0.423 & 0.502 & 0.583 \\
\midrule
\multirow{2}{*}{Relation3D\cite{Lu2025Relation3D}}
& Original & 0.448 & 0.507 & 0.572 & 0.483 & 0.638 & 0.684 & 0.251 & 0.326 & 0.451 & 0.405 & 0.487 & 0.564 \\
& Adaptive & 0.483 & 0.584 & 0.697 & 0.583 & 0.646 & 0.711 & 0.293 & 0.344 & 0.478 & 0.434 & 0.493 & 0.579 \\
\midrule
\multirow{2}{*}{Ours}
& Original & 0.543 & 0.624 & 0.691 & 0.566 & 0.647 & 0.721 & 0.275 & 0.387 & 0.510 & 0.431 & 0.509 & 0.576 \\
& Adaptive & 0.558 & 0.687 & 0.714 & 0.589 & 0.663 & 0.747 & 0.305 & 0.422 & 0.536 & 0.453 & 0.511 & 0.594 \\
\bottomrule
\end{tabular}
\end{table*}

\textbf{Ablation Study on Cross-Scene Transferability.}
Models train on source scenes and are tested on unseen UrbanBIS targets with different building densities, layouts, point-cloud quality, and acquisition conditions. Our model processes complete target scenes; block-based methods use the fixed recovery procedure. Under the same original-scene evaluation, our method leads in AP, AP$_{50}$, and AP$_{25}$ in all three transfer settings (\cref{tab:cross_scene_seg}). This supports cross-scene transferability of the complete framework, although performance remains affected by domain shifts.

\begin{table*}[!t]
\centering
\caption{Ablation Study on Cross-Scene Building Instance Segmentation on UrbanBIS under Scene-Level Evaluation.}
\label{tab:cross_scene_seg}
\setlength{\tabcolsep}{4pt}
\begin{tabular}{>{\centering\arraybackslash}p{0.13\textwidth}|*{3}{>{\centering\arraybackslash}p{0.06\textwidth}}|*{3}{>{\centering\arraybackslash}p{0.06\textwidth}}|*{3}{>{\centering\arraybackslash}p{0.06\textwidth}}}
\toprule
\multirow{2}{*}{Method}
& \multicolumn{3}{c|}{\shortstack{Train: Qingdao + Wuhu\\Test: Longhua}}
& \multicolumn{3}{c|}{\shortstack{Train: Campus\\Test: Qingdao + Wuhu}}
& \multicolumn{3}{c}{\shortstack{Train: Longhua\\Test: Yingrenshi}} \\
& AP & AP$_{50}$ & AP$_{25}$ & AP & AP$_{50}$ & AP$_{25}$ & AP & AP$_{50}$ & AP$_{25}$ \\
\midrule
PointGroup\cite{Jiang2020PointGroup} & 0.214 & 0.273 & 0.419 & 0.224 & 0.324 & 0.465 & 0.515 & \underline{0.624} & 0.629 \\
HAIS\cite{Chen2021HAIS} & 0.164 & 0.257 & 0.389 & 0.359 & \underline{0.501} & 0.561 & 0.434 & 0.522 & 0.609 \\
SoftGroup\cite{Vu2022SoftGroup} & 0.116 & 0.205 & 0.303 & \underline{0.399} & 0.465 & 0.548 & 0.431 & 0.541 & 0.559 \\
DyCo3D\cite{He2021dyco3d} & 0.041 & 0.076 & 0.348 & 0.012 & 0.039 & 0.252 & 0.023 & 0.133 & 0.404 \\
DKNet\cite{wu2022dknet} & 0.133 & 0.216 & 0.293 & 0.081 & 0.129 & 0.205 & 0.304 & 0.397 & 0.401 \\
B-Seg\cite{Yang2023UrbanBIS} & \underline{0.243} & \underline{0.318} & \underline{0.425} & 0.342 & 0.457 & \underline{0.565} & \underline{0.573} & 0.608 & \underline{0.642} \\
Ours & \textbf{0.288} & \textbf{0.384} & \textbf{0.493} & \textbf{0.432} & \textbf{0.574} & \textbf{0.635} & \textbf{0.601} & \textbf{0.663} & \textbf{0.714} \\
\bottomrule
\end{tabular}
\vspace{-0.4cm}
\end{table*}

\textbf{Ablation Study on Unassigned Blocks.}
We compare training with building-aware blocks alone against adding unassigned blocks, which contain points not assigned to any BEV building candidate.
As shown in \cref{tab:unassigned_ablation}, adding unassigned blocks improves instance segmentation performance. This suggests that unassigned blocks help maintain scene coverage and reduce the influence of missed or incomplete BEV building candidates. Without unassigned blocks, the training data is overly concentrated around detected building candidates, which may weaken the model's ability to distinguish buildings from surrounding non-building structures and to handle regions missed by the 2D perception stage.

\begin{table}[!t]
\centering
\caption{Ablation Study on Unassigned Blocks in Adaptive Region Dividing on the Qingdao Scene of UrbanBIS.}
\label{tab:unassigned_ablation}
\begin{tabular}{ccccc}
\toprule
Building blocks & Unassigned blocks & AP & AP$_{50}$ & AP$_{25}$ \\
\midrule
$\checkmark$ &  & 0.409 & 0.587 & 0.650 \\
$\checkmark$ & $\checkmark$ & \textbf{0.558} & \textbf{0.687} & \textbf{0.714} \\
\bottomrule
\end{tabular}
\vspace{-0.5cm}
\end{table}

\subsubsection{Fine-grained Classification}
\textbf{Ablation Study on Training Instance Sources.}
\Cref{tab:gt_pred_cls_ablation} compares training on predicted versus ground-truth instances, with both classifiers evaluated on the same predicted instances. Ground-truth training performs better, consistent with benefits from cleaner supervision despite the train-test input gap; training on predicted instances matches the inference input source but introduces segmentation noise.

\begin{table}[!t]
\centering
\caption{Ablation Study on Training with First-Stage Predicted Instances and Ground-Truth Instances for Fine-Grained Classification.}
\label{tab:gt_pred_cls_ablation}
\begin{tabular}{ccccc}
\toprule
Train inst. & Test inst. & OA & MCA & Macro-F1 \\
\midrule
Predicted & Predicted & 0.671 & 0.438 & 0.402 \\
GT & Predicted & \textbf{0.734} & \textbf{0.516} & \textbf{0.536} \\
\bottomrule
\end{tabular}
\end{table}

\textbf{Ablation Study on Classification and Contrastive Losses.}
\Cref{tab:loss_ablation} compares standard cross-entropy (CE), class-balanced weighted cross-entropy (WCE), supervised contrastive loss (SupCon), and Spatially-Supervised Contrastive Learning (Spatial SupCon). WCE addresses class imbalance; SupCon structures category embeddings; Spatial SupCon gives larger weights to spatially closer same-category, same-scene positives during training.
\begin{table}[!t]
\centering
\caption{Ablation Study on Loss Functions for Fine-Grained Building Function Classification.}
\label{tab:loss_ablation}
\setlength{\tabcolsep}{3.5pt}
\begin{tabular}{ccccccc}
\toprule
CE & WCE & SupCon & Spatial SupCon & OA & MCA & Macro-F1 \\
\midrule
$\checkmark$ &  &  &  & 0.510 & 0.424 & 0.324 \\
 & $\checkmark$ &  &  & 0.717 & 0.434 & 0.448 \\
 & $\checkmark$ & $\checkmark$ &  & 0.723 & 0.440 & 0.452 \\
 & $\checkmark$ &  & $\checkmark$ & \textbf{0.734} & \textbf{0.516} & \textbf{0.536} \\
\bottomrule
\end{tabular}
\vspace{-0.4cm}
\end{table}
The results show that class reweighting and Spatially-Supervised Contrastive Learning improve fine-grained classification. Compared with SupCon, Spatial SupCon increases OA, MCA, and Macro-F1 by $0.011$, $0.076$, and $0.084$, respectively, supporting the effectiveness of spatially informed positive-pair weighting.

\textbf{Ablation Study on the Contrastive Temperature.}
\Cref{tab:temperature_ablation} evaluates the contrastive temperature $\tau$, which controls the sharpness of the contrastive distribution. A smaller $\tau$ emphasizes hard positive and negative pairs, while a larger $\tau$ produces a smoother objective. The results identify an appropriate contrastive temperature for building-level representation learning in the proposed classifier. Among the tested values, $\tau=0.1$ achieves the highest OA, MCA, and Macro-F1, supporting the selected setting.

\begin{table}[!t]
\centering
\caption{Ablation Study on the Temperature Coefficient $\tau$ in Spatially-Supervised Contrastive Loss.}
\label{tab:temperature_ablation}
\begin{tabular}{cccc}
\toprule
$\tau$ & OA & MCA & Macro-F1 \\
\midrule
0.05 & 0.515 & 0.380 & 0.352 \\
0.10 & \textbf{0.734} & \textbf{0.516} & \textbf{0.536} \\
0.20 & 0.490 & 0.368 & 0.353 \\
0.50 & 0.608 & 0.357 & 0.352 \\
\bottomrule
\end{tabular}
\vspace{-0.4cm}
\end{table}

\textbf{Ablation Study on Context Scale.}
\Cref{tab:context_scale_ablation} compares classification without context and with different context ranges. A 10 m expansion achieves the highest OA, MCA, and Macro-F1 among the tested settings; larger ranges offer no further improvement and may include unrelated objects, illustrating the balance between useful context and noise.

\begin{table}[!t]
\centering
\caption{Ablation Study on Context Scale for Fine-Grained Building Function Classification.}
\label{tab:context_scale_ablation}
\begin{tabular}{cccc}
\toprule
Context scale & OA & MCA & Macro-F1 \\
\midrule
No context & 0.623 & 0.339 & 0.348 \\
5 m & 0.555 & 0.396 & 0.377 \\
10 m & \textbf{0.734} & \textbf{0.516} & \textbf{0.536} \\
15 m & 0.512 & 0.323 & 0.334 \\
20 m & 0.598 & 0.360 & 0.359 \\
\bottomrule
\end{tabular}
\end{table}

\section{Discussion and Conclusion}

This work presents a framework for urban building instance segmentation and fine-grained classification. Adaptive region dividing preserves building structures and context during training, while scene-level inference evaluates complete instances in their scene space. The fine-grained classifier combines building bodies with context, uses class-balanced weighted cross-entropy to alleviate long-tailed bias, and employs spatially-supervised contrastive learning to improve feature discrimination. Experiments on UrbanBIS and STPLS3D demonstrate consistent improvements in scene-level segmentation, cross-scene transferability, and building function recognition. The framework depends on the quality of 2D building candidates, and processing large scenes incurs memory costs; segmentation errors may propagate to classification. Future work will investigate 2D--3D candidate refinement, memory-efficient scene-level inference, cross-city adaptation, and multimodal functional recognition.

\small{
\section*{Acknowledgements}
This work was supported by National Key R\&D Program of China (2024YFB3908500, 2024YFB3908504), NSFC (62202312), ICFCRT (W2441020), Shenzhen Science and Technology Program (KJZD20240903100022028, KQTD20210811090044003), Scientific Foundation for Youth Scholars of Shenzhen University, and Scientific Development Fund from Guangdong Provincial Key Laboratory of Visual Media and Multidimensional Intelligence.}

\bibliographystyle{IEEEtran}
\bibliography{main}

@InProceedings{Qi2017PointNet,
  author    = {Qi, Charles R. and Su, Hao and Mo, Kaichun and Guibas, Leonidas J.},
  booktitle = {Proc. IEEE Conference on Computer Vision and Pattern Recognition},
  title     = {{PointNet}: Deep learning on point sets for {3D} classification and segmentation},
  year      = {2017},
  pages     = {652--660},
  series    = {CVPR},
  doi       = {10.1109/CVPR.2017.16},
}

@InProceedings{Qi2017PointNetPlusPlus,
  author    = {Qi, Charles R. and Yi, Li and Su, Hao and Guibas, Leonidas J.},
  booktitle = {Proc. Advances in Neural Information Processing Systems},
  title     = {{PointNet++}: Deep hierarchical feature learning on point sets in a metric space},
  year      = {2017},
  pages     = {5099--5108},
  series    = {NeurIPS},
}

@Article{Wang2019DynamicGraphCNN,
  author  = {Wang, Yue and Sun, Yongbin and Liu, Ziwei and Sarma, Sanjay E. and Bronstein, Michael M. and Solomon, Justin M.},
  journal = {ACM Transactions on Graphics},
  title   = {Dynamic graph {CNN} for learning on point clouds},
  year    = {2019},
  number  = {5},
  pages   = {1--12},
  volume  = {38},
  doi     = {10.1145/3326362},
}

@InProceedings{Zhao2021PointTransformer,
  author    = {Zhao, Hengshuang and Jiang, Li and Jia, Jiaya and Torr, Philip H. S. and Koltun, Vladlen},
  booktitle = {Proc. IEEE/CVF International Conference on Computer Vision},
  title     = {Point transformer},
  year      = {2021},
  pages     = {16259--16268},
  series    = {ICCV},
  doi       = {10.1109/ICCV48922.2021.01595},
}

@InProceedings{Wu2024PointTransformerV3,
  author    = {Wu, Xiaoyang and Jiang, Li and Wang, Peng-Shuai and Liu, Zhijian and Liu, Xihui and Qiao, Yu and Ouyang, Wanli and He, Tong and Zhao, Hengshuang},
  booktitle = {Proc. IEEE/CVF Conference on Computer Vision and Pattern Recognition},
  title     = {Point transformer {V3}: Simpler, faster, stronger},
  year      = {2024},
  pages     = {4840--4851},
  series    = {CVPR},
  doi       = {10.1109/CVPR52733.2024.00463},
}

@InProceedings{Yang2019BoNet,
  author    = {Yang, Bo and Wang, Jianan and Clark, Ronald and Hu, Qingyong and Wang, Sen and Markham, Andrew and Trigoni, Niki},
  booktitle = {Proc. Advances in Neural Information Processing Systems},
  title     = {Learning object bounding boxes for {3D} instance segmentation on point clouds},
  year      = {2019},
  pages     = {6740--6749},
  series    = {NeurIPS},
}

@InProceedings{Jiang2020PointGroup,
  author    = {Jiang, Li and Zhao, Hengshuang and Shi, Shaoshuai and Liu, Shu and Fu, Chi-Wing and Jia, Jiaya},
  booktitle = {Proc. IEEE/CVF Conference on Computer Vision and Pattern Recognition},
  title     = {{PointGroup}: Dual-set point grouping for {3D} instance segmentation},
  year      = {2020},
  pages     = {4867--4876},
  series    = {CVPR},
  doi       = {10.1109/CVPR42600.2020.00492},
}

@InProceedings{Chen2021HAIS,
  author    = {Chen, Shaoyu and Fang, Jiemin and Zhang, Qian and Liu, Wenyu and Wang, Xinggang},
  booktitle = {Proc. IEEE/CVF International Conference on Computer Vision},
  title     = {Hierarchical aggregation for {3D} instance segmentation},
  year      = {2021},
  pages     = {15467--15476},
  series    = {ICCV},
  doi       = {10.1109/ICCV48922.2021.01518},
}

@InProceedings{Vu2022SoftGroup,
  author    = {Vu, Thang and Kim, Kookhoi and Luu, Tung M. and Nguyen, Xuan Thanh and Yoo, Chang D.},
  booktitle = {Proc. IEEE/CVF Conference on Computer Vision and Pattern Recognition},
  title     = {{SoftGroup} for {3D} instance segmentation on point clouds},
  year      = {2022},
  pages     = {2708--2717},
  series    = {CVPR},
  doi       = {10.1109/CVPR52688.2022.00273},
}

@InProceedings{Ngo2023ISBNet,
  author    = {Ngo, Tuan Duc and Hua, Binh-Son and Nguyen, Khoi},
  booktitle = {Proc. IEEE/CVF Conference on Computer Vision and Pattern Recognition},
  title     = {{ISBNet}: A {3D} point cloud instance segmentation network with instance-aware sampling and box-aware dynamic convolution},
  year      = {2023},
  pages     = {13550--13559},
  series    = {CVPR},
  doi       = {10.1109/CVPR52729.2023.01302},
}

@InProceedings{Schult2023Mask3D,
  author    = {Schult, Jonas and Engelmann, Francis and Hermans, Alexander and Litany, Or and Tang, Siyu and Leibe, Bastian},
  booktitle = {Proc. IEEE International Conference on Robotics and Automation},
  title     = {{Mask3D}: Mask transformer for {3D} semantic instance segmentation},
  year      = {2023},
  pages     = {8216--8223},
  series    = {ICRA},
  doi       = {10.1109/ICRA48891.2023.10160590},
}

@InProceedings{Kolodiazhnyi2024OneFormer3D,
  author    = {Kolodiazhnyi, Maksim and Vorontsova, Anna and Konushin, Anton and Rukhovich, Danila},
  booktitle = {Proc. IEEE/CVF Conference on Computer Vision and Pattern Recognition},
  title     = {{OneFormer3D}: One transformer for unified point cloud segmentation},
  year      = {2024},
  pages     = {20943--20953},
  series    = {CVPR},
  doi       = {10.1109/CVPR52733.2024.01979},
}

@InProceedings{Lu2025Relation3D,
  author    = {Lu, Jiahao and Deng, Jiacheng},
  booktitle = {Proc. IEEE/CVF Conference on Computer Vision and Pattern Recognition},
  title     = {{Relation3D}: Enhancing relation modeling for point cloud instance segmentation},
  year      = {2025},
  pages     = {8889--8899},
  series    = {CVPR},
  doi       = {10.1109/CVPR52734.2025.00831},
}

@InProceedings{Carion2020DETR,
  author    = {Carion, Nicolas and Massa, Francisco and Synnaeve, Gabriel and Usunier, Nicolas and Kirillov, Alexander and Zagoruyko, Sergey},
  booktitle = {Proc. European Conference on Computer Vision},
  title     = {End-to-end object detection with transformers},
  year      = {2020},
  pages     = {213--229},
  series    = {ECCV},
  doi       = {10.1007/978-3-030-58452-8_13},
}

@InProceedings{Cheng2022Mask2Former,
  author    = {Cheng, Bowen and Misra, Ishan and Schwing, Alexander G. and Kirillov, Alexander and Girdhar, Rohit},
  booktitle = {Proc. IEEE/CVF Conference on Computer Vision and Pattern Recognition},
  title     = {{Masked-attention Mask Transformer} for universal image segmentation},
  year      = {2022},
  pages     = {1290--1299},
  series    = {CVPR},
  doi       = {10.1109/CVPR52688.2022.00135},
}

@Article{Kuhn1955HungarianMethod,
  author  = {Kuhn, Harold W.},
  journal = {Naval Research Logistics Quarterly},
  title   = {The {Hungarian} method for the assignment problem},
  year    = {1955},
  number  = {1--2},
  pages   = {83--97},
  volume  = {2},
  doi     = {10.1002/nav.3800020109},
}

@InProceedings{kirillov2023segany,
  author    = {Kirillov, Alexander and Mintun, Eric and Ravi, Nikhila and Mao, Hanzi and Rolland, Chloe and Gustafson, Laura and Xiao, Tete and Whitehead, Spencer and Berg, Alexander C. and Lo, Wan-Yen and Doll{\'a}r, Piotr and Girshick, Ross},
  booktitle = {Proc. IEEE/CVF International Conference on Computer Vision},
  title     = {Segment Anything},
  year      = {2023},
  pages     = {4015--4026},
  series    = {ICCV},
  doi       = {10.1109/ICCV51070.2023.00371},
}

@InProceedings{liu2023grounding,
  author    = {Liu, Shilong and Zeng, Zhaoyang and Ren, Tianhe and Li, Feng and Zhang, Hao and Yang, Jie and Jiang, Qing and Li, Chunyuan and Yang, Jianwei and Su, Hang and Zhu, Jun and Zhang, Lei},
  booktitle = {Proc. European Conference on Computer Vision},
  title     = {Grounding {DINO}: Marrying {DINO} with grounded pre-training for open-set object detection},
  year      = {2024},
  pages     = {38--55},
  series    = {ECCV},
  doi       = {10.1007/978-3-031-72970-6_3},
}

@InProceedings{Ravi2024SAM2,
  author    = {Ravi, Nikhila and Gabeur, Valentin and Hu, Yuan-Ting and Hu, Ronghang and Ryali, Chaitanya and Ma, Tengyu and Khedr, Haitham and R{\"a}dle, Roman and Rolland, Chloe and Gustafson, Laura and Mintun, Eric and Pan, Junting and Alwala, Kalyan Vasudev and Carion, Nicolas and Wu, Chao-Yuan and Girshick, Ross and Doll{\'a}r, Piotr and Feichtenhofer, Christoph},
  booktitle = {Proc. International Conference on Learning Representations},
  title     = {{SAM 2}: Segment anything in images and videos},
  year      = {2025},
  series    = {ICLR},
  url       = {https://openreview.net/forum?id=Ha6RTeWMd0},
}

@Misc{ren2024grounded,
  author        = {Ren, Tianhe and Liu, Shilong and Zeng, Ailing and Lin, Jing and Li, Kunchang and Cao, He and Chen, Jiayu and Huang, Xinyu and Chen, Yukang and Yan, Feng and Zeng, Zhaoyang and Zhang, Hao and Li, Feng and Yang, Jie and Li, Hongyang and Jiang, Qing and Zhang, Lei},
  title         = {Grounded {SAM}: Assembling open-world models for diverse visual tasks},
  year          = {2024},
  eprint        = {2401.14159},
  archiveprefix = {arXiv},
  primaryclass  = {cs.CV},
  url           = {https://arxiv.org/abs/2401.14159},
}

@Article{Chen2024RSPrompter,
  author  = {Chen, Keyan and Liu, Chenyang and Chen, Hao and Zhang, Haotian and Li, Wenyuan and Zou, Zhengxia and Shi, Zhenwei},
  journal = {IEEE Transactions on Geoscience and Remote Sensing},
  title   = {{RSPrompter}: Learning to prompt for remote sensing instance segmentation based on visual foundation model},
  year    = {2024},
  pages   = {1--17},
  volume  = {62},
  doi     = {10.1109/TGRS.2024.3356074},
}

@Article{Zhou2024MeSAM,
  author  = {Zhou, Xichuan and Liang, Fu and Chen, Lihui and Liu, Haijun and Song, Qianqian and Vivone, Gemine and Chanussot, Jocelyn},
  journal = {IEEE Transactions on Geoscience and Remote Sensing},
  title   = {{MeSAM}: Multiscale enhanced Segment Anything Model for optical remote sensing images},
  year    = {2024},
  pages   = {1--15},
  volume  = {62},
  doi     = {10.1109/TGRS.2024.3398038},
}

@InProceedings{Wang2023SAMRS,
  author    = {Wang, Di and Zhang, Jing and Du, Bo and Xu, Minqiang and Liu, Lin and Tao, Dacheng and Zhang, Liangpei},
  booktitle = {Advances in Neural Information Processing Systems},
  title     = {{SAMRS}: Scaling-up remote sensing segmentation dataset with Segment Anything Model},
  year      = {2023},
  pages     = {8815--8827},
  volume    = {36},
  doi       = {10.52202/075280-0385},
}

@InProceedings{Dai2017ScanNet,
  author    = {Dai, Angela and Chang, Angel X. and Savva, Manolis and Halber, Maciej and Funkhouser, Thomas and Nie{\ss}ner, Matthias},
  booktitle = {Proc. IEEE Conference on Computer Vision and Pattern Recognition},
  title     = {{ScanNet}: Richly-annotated {3D} reconstructions of indoor scenes},
  year      = {2017},
  pages     = {5828--5839},
  series    = {CVPR},
  doi       = {10.1109/CVPR.2017.261},
}

@InProceedings{Chen2022STPLS3D,
  author    = {Chen, Meida and Hu, Qingyong and Yu, Zifan and Thomas, Hugues and Feng, Andrew and Hou, Yu and McCullough, Kyle and Ren, Fengbo and Soibelman, Lucio},
  booktitle = {Proc. British Machine Vision Conference},
  title     = {{STPLS3D}: A large-scale synthetic and real aerial photogrammetry {3D} point cloud dataset},
  year      = {2022},
  series    = {BMVC},
  url       = {https://www.stpls3d.com/},
}

@InProceedings{Yang2023UrbanBIS,
  author    = {Yang, Guoqing and Xue, Fuyou and Zhang, Qi and Xie, Ke and Fu, Chi-Wing and Huang, Hui},
  booktitle = {Proc. ACM SIGGRAPH Conference Proceedings},
  title     = {{UrbanBIS}: A large-scale benchmark for fine-grained urban building instance segmentation},
  year      = {2023},
  series    = {SIGGRAPH},
  doi       = {10.1145/3588432.3591508},
}

@InProceedings{Lin2015BilinearCNN,
  author    = {Lin, Tsung-Yu and RoyChowdhury, Aruni and Maji, Subhransu},
  booktitle = {Proc. IEEE International Conference on Computer Vision},
  title     = {Bilinear {CNN} models for fine-grained visual recognition},
  year      = {2015},
  pages     = {1449--1457},
  series    = {ICCV},
  doi       = {10.1109/ICCV.2015.170},
}

@InProceedings{Khosla2020SupervisedContrastive,
  author    = {Khosla, Prannay and Teterwak, Piotr and Wang, Chen and Sarna, Aaron and Tian, Yonglong and Isola, Phillip and Maschinot, Aaron and Liu, Ce and Krishnan, Dilip},
  booktitle = {Proc. Advances in Neural Information Processing Systems},
  title     = {Supervised contrastive learning},
  year      = {2020},
  pages     = {18661--18673},
  series    = {NeurIPS},
}

@InProceedings{Cui2019ClassBalanced,
  author    = {Cui, Yin and Jia, Menglin and Lin, Tsung-Yi and Song, Yang and Belongie, Serge},
  booktitle = {Proc. IEEE/CVF Conference on Computer Vision and Pattern Recognition},
  title     = {Class-balanced loss based on effective number of samples},
  year      = {2019},
  pages     = {9260--9269},
  series    = {CVPR},
  doi       = {10.1109/CVPR.2019.00949},
}

@InProceedings{Uy2019ScanObjectNN,
  author    = {Uy, Mikaela Angelina and Pham, Quang-Hieu and Hua, Binh-Son and Nguyen, Thanh and Yeung, Sai-Kit},
  booktitle = {Proc. IEEE/CVF International Conference on Computer Vision},
  title     = {Revisiting point cloud classification: A new benchmark dataset and classification model on real-world data},
  year      = {2019},
  pages     = {1588--1597},
  series    = {ICCV},
  doi       = {10.1109/ICCV.2019.00167},
}

@Article{Zhou2023BuildingUse,
  author  = {Zhou, Wen and Persello, Claudio and Li, Mengmeng and Stein, Alfred},
  journal = {Remote Sensing of Environment},
  title   = {Building use and mixed-use classification with a transformer-based network fusing satellite images and geospatial textual information},
  year    = {2023},
  pages   = {113767},
  volume  = {297},
  doi     = {10.1016/j.rse.2023.113767},
}

@Article{Kang2018BuildingInstanceClassification,
  author  = {Kang, Jian and K{\"o}rner, Marco and Wang, Yuanyuan and Taubenb{\"o}ck, Hannes and Zhu, Xiao Xiang},
  journal = {ISPRS Journal of Photogrammetry and Remote Sensing},
  title   = {Building instance classification using street view images},
  year    = {2018},
  pages   = {44--59},
  volume  = {145},
  doi     = {10.1016/j.isprsjprs.2018.02.006},
}

@Article{Wang2021BuildingFunctionMapping,
  author  = {Wang, Jionghua and Luo, Haowen and Li, Wenyu and Huang, Bo},
  journal = {Remote Sensing},
  title   = {Building function mapping using multisource geospatial big data: A case study in Shenzhen, China},
  year    = {2021},
  number  = {23},
  pages   = {4751},
  volume  = {13},
  doi     = {10.3390/rs13234751},
}

@InProceedings{Ma2022PointMLP,
  author    = {Ma, Xu and Qin, Can and You, Haoxuan and Ran, Haoxi and Fu, Yun},
  booktitle = {Proc. International Conference on Learning Representations},
  title     = {Rethinking network design and local geometry in point cloud: A simple residual {MLP} framework},
  year      = {2022},
  series    = {ICLR},
}

@InProceedings{Qian2022PointNeXt,
  author    = {Qian, Guocheng and Li, Yuchen and Peng, Houwen and Mai, Jinjie and Hammoud, Hasan and Elhoseiny, Mohamed and Ghanem, Bernard},
  booktitle = {Proc. Advances in Neural Information Processing Systems},
  title     = {{PointNeXt}: Revisiting {PointNet++} with improved training and scaling strategies},
  year      = {2022},
  pages     = {23192--23204},
  series    = {NeurIPS},
}

@InProceedings{Liang2024PointMamba,
  author    = {Liang, Dingkang and Zhou, Xin and Xu, Wei and Zhu, Xingkui and Zou, Zhikang and Ye, Xiaoqing and Tan, Xiao and Bai, Xiang},
  booktitle = {Proc. Advances in Neural Information Processing Systems},
  title     = {{PointMamba}: A simple state space model for point cloud analysis},
  year      = {2024},
  pages     = {32653--32677},
  series    = {NeurIPS},
  doi       = {10.52202/079017-1026},
}

@InProceedings{Shi2025PointKAN,
  author    = {Shi, Yan and He, Qingdong and Liu, Yijun and Su, Jingyong},
  booktitle = {Proc. International Conference on Multimedia Retrieval},
  title     = {{KAN} or {MLP}? Point cloud shows the way forward},
  year      = {2026},
  pages     = {1375--1384},
  series    = {ICMR},
  doi       = {10.1145/3805622.3810696},
}

@InProceedings{He2021dyco3d,
  author    = {He, Tong and Shen, Chunhua and van den Hengel, Anton},
  booktitle = {Proc. IEEE/CVF Conference on Computer Vision and Pattern Recognition},
  title     = {{DyCo3D}: Robust instance segmentation of {3D} point clouds through dynamic convolution},
  year      = {2021},
  pages     = {354--363},
  series    = {CVPR},
  doi       = {10.1109/CVPR46437.2021.00042},
}

@InProceedings{wu2022dknet,
  author    = {Wu, Yizheng and Shi, Min and Du, Shuaiyuan and Lu, Hao and Cao, Zhiguo and Zhong, Weicai},
  booktitle = {Proc. European Conference on Computer Vision},
  title     = {{3D} instances as {1D} kernels},
  year      = {2022},
  pages     = {235--252},
  series    = {ECCV},
  doi       = {10.1007/978-3-031-19818-2_14},
}

@InProceedings{lai2023maft,
  author    = {Lai, Xin and Yuan, Yuhui and Chu, Ruihang and Chen, Yukang and Hu, Han and Jia, Jiaya},
  booktitle = {Proc. IEEE/CVF International Conference on Computer Vision},
  title     = {Mask-attention-free transformer for {3D} instance segmentation},
  year      = {2023},
  pages     = {3693--3703},
  series    = {ICCV},
  doi       = {10.1109/ICCV51070.2023.00342},
}

@InProceedings{shin2024spherical,
  author    = {Shin, Sangyun and Zhou, Kaichen and Vankadari, Madhu and Markham, Andrew and Trigoni, Niki},
  booktitle = {Proc. IEEE/CVF Conference on Computer Vision and Pattern Recognition},
  title     = {Spherical Mask: Coarse-to-fine {3D} point cloud instance segmentation with spherical representation},
  year      = {2024},
  pages     = {4060--4069},
  series    = {CVPR},
  doi       = {10.1109/CVPR52733.2024.00389},
}

@InProceedings{td3d,
  author    = {Kolodiazhnyi, Maksim and Vorontsova, Anna and Konushin, Anton and Rukhovich, Danila},
  booktitle = {Proc. IEEE/CVF Winter Conference on Applications of Computer Vision},
  title     = {Top-down beats bottom-up in {3D} instance segmentation},
  year      = {2024},
  pages     = {3554--3562},
  series    = {WACV},
  doi       = {10.1109/WACV57701.2024.00353},
}

@InProceedings{roh2024EASE,
  author    = {Roh, Wonseok and Jung, Hwanhee and Nam, Giljoo and Yeom, Jinseop and Park, Hyunje and Yoon, Sang Ho and Kim, Sangpil},
  booktitle = {Proc. IEEE/CVF Conference on Computer Vision and Pattern Recognition},
  title     = {Edge-aware {3D} instance segmentation network with intelligent semantic prior},
  year      = {2024},
  pages     = {20644--20653},
  series    = {CVPR},
  doi       = {10.1109/CVPR52733.2024.01951},
}

@InProceedings{roh2025IKNE,
  author    = {Roh, Wonseok and Jung, Hwanhee and Nam, Giljoo and Lee, Dong In and Park, Hyeongcheol and Yoon, Sang Ho and Joo, Jungseock and Kim, Sangpil},
  booktitle = {Proc. IEEE/CVF Conference on Computer Vision and Pattern Recognition},
  title     = {Insightful instance features for {3D} instance segmentation},
  year      = {2025},
  pages     = {14057--14067},
  series    = {CVPR},
  doi       = {10.1109/CVPR52734.2025.01312},
}

@Article{Wu2024IntraCrossModal,
  author = {Wu, Yue and Liu, Jiaming and Gong, Maoguo and Gong, Peiran and Fan, Xiaolong and Qin, A. K. and Miao, Qiguang and Ma, Wenping},
  title = {Self-Supervised Intra-Modal and Cross-Modal Contrastive Learning for Point Cloud Understanding},
  journal = {IEEE Transactions on Multimedia},
  year = {2024},
  volume = {26},
  pages = {1626--1638},
  doi = {10.1109/tmm.2023.3284591},
}

@Article{Sun2025Oversegmentation,
  author = {Sun, Yifan and Dai, Chenguang and Li, Wenke and Zhang, Yongsheng and Ji, Song and Yu, Anzhu and Chen, Yiping and Wang, Hanyun},
  title = {Supervised Contrastive Learning for Indoor Point Cloud Oversegmentation},
  journal = {IEEE Transactions on Multimedia},
  year = {2025},
  volume = {27},
  pages = {9189--9201},
  doi = {10.1109/tmm.2025.3613177},
}

@Article{Chen2026Ambiguity,
  author = {Chen, Yang and Duan, Yueqi and Sun, Haowen and Lu, Jiwen and Tan, Yap-Peng},
  title = {Ambiguity-Aware Point Cloud Segmentation by Adaptive Margin Contrastive Learning},
  journal = {IEEE Transactions on Multimedia},
  year = {2026},
  volume = {28},
  pages = {439--453},
  doi = {10.1109/tmm.2025.3623494},
}

\end{document}